\documentclass[letterpaper, 10 pt, conference]{ieeeconf}  

\IEEEoverridecommandlockouts         
\usepackage{graphicx} 
\usepackage[utf8]{inputenc}
\usepackage{cite}
\usepackage{url}
\usepackage{color}
\usepackage{xspace}
\usepackage{algorithm}
\usepackage[noend]{algpseudocode}
\usepackage{siunitx}
\usepackage{midfloat}
\usepackage{subcaption}
\usepackage{multirow}
\usepackage{fancyhdr}

\newcommand{\stt}[1]{{\small\texttt{#1}}}

\title{\LARGE \bf
Autonomous tissue retraction with a biomechanically informed logic based framework
}

\author{Daniele Meli$^{1*}$, Eleonora Tagliabue$^{1*}$, Diego Dall'Alba$^{1}$, Paolo Fiorini$^{1}$
\thanks{$^{1}$Authors are with Department of Computer Science, University of Verona, Verona, Italy {\tt\small daniele.meli@univr.it}}%
\thanks{*Authors contributed equally and are reported in alphabetical order.}
\thanks{This project has received funding from the European Research Council (ERC) under the European Union's Horizon 2020 research and innovation programme (grant agreement No. 742671 "ARS").}%
}

\begin{document}

\maketitle
\thispagestyle{empty}
\pagestyle{empty}

\thispagestyle{fancy}
\renewcommand{\headrulewidth}{0pt}
\renewcommand{\footrulewidth}{0.5pt}
\cfoot{\footnotesize{\copyright 2021 IEEE. Personal use of this material is permitted. Permission from IEEE must be obtained for all other uses, in any current or future media, including reprinting/republishing this material for advertising or promotional purposes, creating new collective works, for resale or redistribution to servers or lists, or reuse of any copyrighted component of this work in other works.}}

\begin{abstract}
Autonomy in robot-assisted surgery is essential to reduce surgeons' cognitive load and eventually improve the overall surgical outcome. A key requirement for autonomy in a safety-critical scenario as surgery lies in the generation of interpretable plans that rely on expert knowledge. Moreover, the Autonomous Robotic Surgical System (ARSS) must be able to reason on the dynamic and unpredictable anatomical environment, and quickly adapt the surgical plan in case of unexpected situations. In this paper, we present a modular Framework for Robot-Assisted Surgery (FRAS) in deformable anatomical environments. Our framework integrates a logic module for task-level interpretable reasoning, a biomechanical simulation that complements data from real sensors, and a situation awareness module for context interpretation. The framework performance is evaluated on simulated soft tissue retraction, a common surgical task to remove the tissue hiding a region of interest. Results show that the framework has the adaptability required to successfully accomplish the task, handling dynamic environmental conditions and possible failures, while guaranteeing the computational efficiency required in a real surgical scenario. The framework is made publicly available.
\end{abstract}

\begin{keywords}
Autonomous Robotic Surgery; Logic Programming; Biomechanical simulation;
\end{keywords}

\section{INTRODUCTION}
One frontier of research in surgical robotics is the achievement of higher levels of autonomy, which has the potential to improve the quality of an intervention \cite{yang2017medical}. 
In particular, level 2 of autonomy as of \cite{yang2017medical} will introduce the autonomous execution of surgical sub-tasks, limiting human errors and the fatigue of surgeons, who shall monitor the overall execution and focus only on the most critical steps.
An Autonomous Robotic Surgical System (ARSS) must guarantee an \emph{interpretable} behavior, i.e. the input and output to the autonomous system must be easily understandable by a monitoring surgeon, for safety and reliability \cite{fiazza2021design,EURLEX}. Moreover, the surgical scenario is affected by high variability (e.g., patient's anatomy and tissue properties). This requires continuous reasoning on intra-operative data from sensors, in order to responsively adapt to the quick changes of the environment and to recover from possible failures.
\begin{figure}[t]
\centering
\includegraphics[width=.8\linewidth]{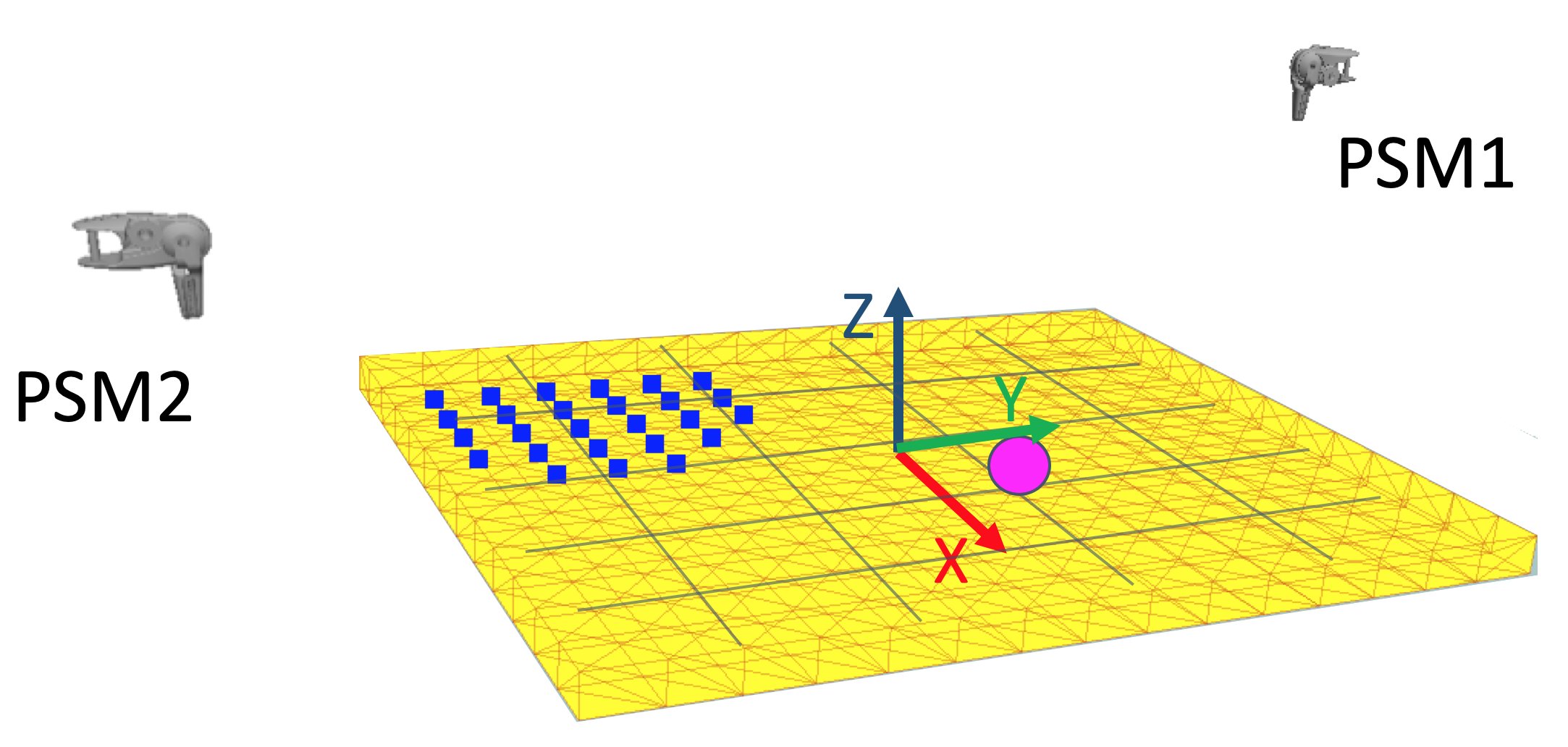}
\caption{The simulated environment considered for the tissue retraction task: deformable tissue (yellow), ROI (magenta sphere), attachment points (blue dots) and common reference frame. Gray lines delineate the $5 \times 5$ discretization for selection of grasping points.}  
\label{fig:setup}
\end{figure}

In this paper, we approach the problem of reaching autonomy level 2 \cite{yang2017medical} in a surgical deformable environment, with a modular framework that integrates interpretable task planning and real-time situation awareness. Interpretability is guaranteed by both encoding prior expert knowledge in logic formalism at task level, and providing semantic interpretation of data from sensors with situation awareness.
The proposed framework builds on our previous work \cite{ginesi2020autonomous}, which exhibits fast computational performance compatible with the surgical time constraints, but has been tested only on a standard surgical training task in rigid domain. 
In this work, we consider a more realistic and significant surgical use case, consisting in Tissue Retraction (TR) to expose a region of interest (e.g., a tumor or other pathological areas). We extend the existing framework to operate in deformable environments by introducing a patient-specific biomechanical simulation that follows the actual condition throughout surgical task execution. The physics-based model estimates the applied forces, allowing to reduce the risk of tissue damage due to high solicitation.
The capabilities of our framework are validated on a simulated TR, executed with the da Vinci Research Kit (dVRK)\cite{kazanzides2014open}.
The specific contributions of this work are the following:
\begin{enumerate}
	\item 
	we show that our Framework for Robot-Assisted Surgery (FRAS) can adapt to the dynamic environmental conditions and re-plan when critical situations occur. To the best of our knowledge, this is the first attempt to failure recovery during TR;
	\item we show that the integration of a biomechanical simulation allows to deal with specific problems of real surgery, e.g. lack of force feedback;
	\item we show that FRAS generates interpretable plans (see attached video), guaranteeing computational efficiency even when operating in a deformable environment;
	\item we publicly share the framework at \url{https://gitlab.com/altairLab/tissue_retraction/-/tree/v1.0}.
\end{enumerate}
The paper is organized as follows: in Section \ref{sec:sota} we review state of the art in autonomous TR, and we motivate the use of logic programming; in Section \ref{sec:method} we describe the TR task and the proposed autonomous framework; Section \ref{sec:exp} reports the experimental results of this paper, and Section \ref{sec:conclusion} summarizes the main achievements and possible future research.

\section{RELATED WORKS}
\label{sec:sota}
Recent research has investigated the automation of basic operations in surgery, e.g. suturing \cite{pedram2020autonomous}, needle insertion \cite{muradore2015development} and dissection \cite{nagy2019dvrk}. 
Autonomous TR combining standard motion planning algorithms with features for optimal grasping identification is performed in \cite{attanasio2020autonomous}.
In \cite{patil2010toward,jansen2009surgical}, authors have focused on the generation of optimal robotic trajectories and grasping point selection for TR, maximizing target visibility and minimizing tissue damage based on biomechanical simulation.
These works assume the execution workflow is predefined, and do not deal with failure recovery and unconventionial situations which occur in real (surgical) robotic tasks.

Learning-based methods have recently shown the potential to capture variability in the surgical scenario.
In \cite{tagliabue2020soft}, deep reinforcement learning is proposed to learn a policy for robotic TR in a simulated environment.
\cite{shin2019autonomous} shows that efficient transfer of learned policies from simulation to the real system in uncertain environments can be obtained with the incorporation of few human demonstrations. \cite{pedram2019toward} presents a learning strategy which incorporates human knowledge for the definition of task-specific features. However, learning-based methods fail to generalize to situations not seen at training time and do not guarantee any interpretability \cite{goebel2018explainable}, which is essential for safety-critical scenarios as surgery. 

Interpretability for autonomous robots can be guaranteed by \emph{logic-based reasoning systems} \cite{Tenorth}.
They encode prior concepts and specifications about a robotic domain in a knowledge base (e.g. an ontology as proposed by \cite{Tenorth}), using logic formalism which is easily readable by humans.
Planning languages as PDDL then allow to apply logic inference on knowledge \cite{Fox2003}.
A logic-based task planner does not encode pre-defined workflows (as finite state machines do); instead, it performs human-like logic inference from knowledge and data from sensors, in order to generate an appropriate workflow for the current situation. Hence, the logic-based approach formally guarantees that specifications encoded in prior knowledge (which may be, e.g., safety constraints in surgery) are satisfied during the execution.
However, \cite{ICAR19} evidences the limits of ontology-based reasoning, which can only represent a static scenario, while real-time knowledge revision for re-planning is computationally inefficient. For this reason, ontologies are mostly used for supervision and assistance in surgery \cite{nakawala2019deep, katic2013ontology}.
In this paper, we instead use non-monotonic logic programming \cite{dimopoulos1997encoding}, which combines knowledge representation and reasoning, and permits online revision of incomplete or dynamic information.
Although non-monotonic logic programming has never been applied to surgery, examples in other fields \cite{GOSRR18,FFSTT18} show its feasibility in safety-critical scenarios. 
In this work, we rely on \emph{Answer Set Programming (ASP)} \cite{lifschitz2002answer}, a non-monotonic logic paradigm which is more computationally efficient and expressive if compared to its main competitor Prolog \cite{ColmA90}.

\section{METHODS}
\label{sec:method}
In this Section, we describe and motivate the choice of the TR task as use case, and describe the proposed framework for ARSS.
\subsection{Tissue retraction task}
\label{sec:task}
Tissue retraction (TR) is a common task in surgery. 
It aims at grasping and lifting an anatomical layer (e.g., adipose tissue) to reveal and make accessible an underlying region of interest (ROI) (e.g., a tumor).
We assume that the 3D geometric model of the tissue and the 3D position of the ROI to expose are known from pre-operative data (Fig.~\ref{fig:setup}). The task is executed by the two patient-side manipulators (PSMs) of the dVRK.
A challenge of TR is the presence of \emph{attachment points (APs)} between the tissue and the surrounding anatomy, which can be initialized based on statistical information \cite{plantefeve2016patient,nikolaev2020estimation}.
Avoiding APs is important in TR to prevent the application of excessive forces on highly constrained areas, thus reducing the risk of tissue tearing and/or rupture. 
For an efficient execution of TR, we choose the grasping point which maximizes the distance from known APs, while minimizing the distance from ROI for faster exposure. 

The selected grasping point is considered \emph{reachable} by one of the PSMs, depending on its position with respect to the tissue $x-$axis of symmetry (hence, distance on $y-$axis is considered, Fig.~\ref{fig:setup}). 
After grasping, the PSM pulls the tissue up to expose ROI \cite{bhayani2008vinci}. However, other strategies are needed in case the ROI is not exposed and/or critical conditions rise, e.g. excessive applied force, that can be recognized thanks to the feedback provided by a biomechanical simulation.
The reasoner then generates an alternative strategy, e.g. lateral motion to expose the ROI (Fig.~\ref{fig:sceneB}) or grasp re-planning (Fig.~\ref{fig:sceneC}).

\subsection{Framework for Robot Assisted Surgery (FRAS)}
\begin{figure}[t]
\centering
\includegraphics[trim={0 0 0 0},clip,width=0.48\textwidth] {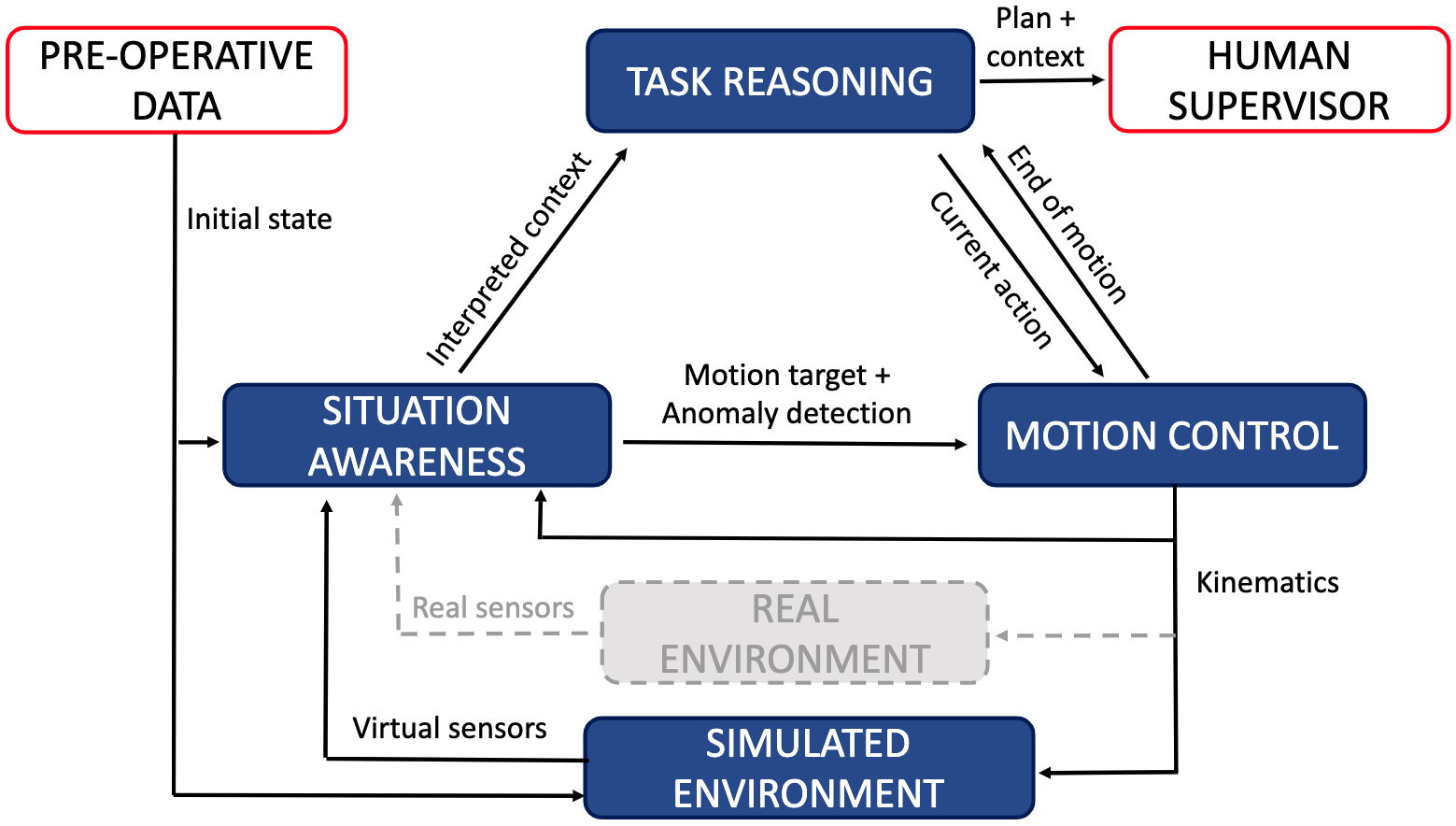}
\caption{Block diagram of the proposed FRAS. Blue blocks represent modules used in this paper, with arrows specifying exchanged information. Grey parts of the scheme allow extension towards application in a real surgical setup.}
\label{fig:framework}
\end{figure}

Fig.~\ref{fig:framework} depicts the main modules of our framework.
FRAS is initialized with pre-operative information, that define the starting anatomical condition and the position of the target to reach. During task execution, a \emph{situation awareness (SA)} module receives information from sensors and dynamic feedback from a biomechanical simulation of the environment. This information is translated to a semantic interpretation (\emph{context}) of the environment state, and sent to the \emph{task reasoner}. The task reasoner generates a plan, i.e. a sequence of actions towards the goal of the task. The actions are communicated to a \emph{motion control (MC)} module, which translates them into motion commands for the robotic arms.
The context and the plan can be monitored by an expert human supervisor, to guarantee the reliability of the ARSS.

\subsubsection{Task reasoner}
This module encodes task knowledge in the logic programming formalism of Answer Set Programming (ASP) \cite{asp_syntax}. 
An ASP program represents the domain of the task with a \emph{description} $\mathcal{D}$ and a \emph{history} $\mathcal{H}$. $\mathcal{D}$ comprises a \emph{signature} $\Sigma$ and axioms. $\Sigma$ is the alphabet of the task, defining relevant attributes of the domain. Attributes may be \emph{statics}, i.e., domain attributes whose values do not change over time; \emph{fluents}, i.e., domain attributes whose values can be changed; and \textit{actions}.
Attributes may be \emph{terms}; \emph{atoms}, i.e. predicates of terms (e.g. \stt{Atom(T$_1$, ..., T$_n$)} is an atom with terms \stt{T$_{1, ..., n}$} as arguments); and their logical negations.
Values of terms are \emph{constants} (either integers or strings).
A term whose value is assigned is \emph{ground}, and an atom is ground if its terms are ground.
Axioms are logical relations between attributes. In this paper, we consider causal rules \stt{A$_h$ :- A$_{b1}$, ..., A$_{bn}$}, defining pre-conditions \stt{A$_{b1, ..., bn}$} (\emph{body} of the rule) for grounding the \emph{head} \stt{A$_h$}.
Following the definition of the task introduced in Sec. \ref{sec:task}, statics are \stt{block} (from now on \stt{B}), representing a generic candidate grasping point on the tissue; and \stt{arm} (from now on \stt{A}), representing the robotic arm (either \stt{psm1} or \stt{psm2}). Fluents are relevant environmental conditions for the task. We define them as atoms with the variable \stt{t} as argument, representing a discrete time step for temporal reasoning.
They include \stt{reachable(A, B, t)} (\stt{B} reachable by \stt{A}), \stt{in\_hand(A, B, t)}, \stt{closed\_gripper(A, t)} (gripper state), \stt{distance(B1, B2, X, t)} (the distance between \stt{B1}-\stt{B2} is \stt{X}), \stt{visible\_ROI(t)}, \stt{max\_height(A, t)} (pulling is no longer possible if excessive force is measured), \stt{fixed(B, t)} (\stt{B} is close to APs), \stt{at(A, B, t)} (location of \stt{A} on tissue) and \stt{above\_ROI(B, t)} (location of ROI under tissue).
All fluents except \stt{distance} are defined as \emph{external} atoms, so they can be grounded from the SA module.

Actions for our task are \stt{reach(A, B, t)}, \stt{grasp(A, B, t)}, \stt{release(A, t)}, \stt{pull(A, B, t)} and \stt{move(A, B, t)} (for lateral motion).
Axioms express logical relations between atoms. For the task of TR, we consider axioms representing causal laws, executability constraints , choice rules and optimization statements. 
Causal laws define \emph{effects of actions} on fluents (e.g., \stt{A}'s gripper is closed after grasping, \stt{closed\_gripper(A, t) :- grasp(A, B, t-1)}). 
Executability constraints specify the set of actions and fluents that cannot hold concurrently (e.g., it is not possible to pull up tissue if the force applied to the tissue exceeds constraints, \stt{:- pull(A, B, t), max\_height(A, t)}).
Choice rules specify the environmental \emph{pre-conditions} which must be satisfied to execute actions. For our task, at most one single action can be executed at each time step. Then, we define a choice rule gathering pre-conditions for actions as \stt{0 \{action(t) : pre-cond(t)\} 1}, which constrains the cardinality of the set of possible actions returned at each time step between 0 and 1. 
Optimization statements define preference in the set of possible actions, in particular for grasping point selection as defined in Sec.~\ref{sec:task}. For instance, to maximize distance between grasping point and APs, we state \stt{\#maximize\{X : reach(A, B1, t), distance(B1, B2, X, t), fixed(B2, t)\}}.
We finally define the goal of the TR task as the constraint that the ROI must be eventually visible within a predefined percentage (\stt{:- visible\_ROI(t)}). 

Given the ASP program, an ASP solver grounds information from sensors and axioms at each time step, and returns them as $\mathcal{H}$, enhancing interpretability of the behavior of ARSS. For task planning, we select from $\mathcal{H}$ only atoms representing actions to command the robot (see attached video).

\subsubsection{Situation awareness}
The SA module continuously monitors the state of the robot and the environment, to provide high-level interpretation of the environmental context and reduce the risk of tissue damage.
The functions of the SA module are described in Alg.~\ref{alg:sa}.
\begin{algorithm}[t]
    \caption{Situation awareness}\label{alg:sa}
    \begin{algorithmic}[1]
        \State \textbf{Input}: $S_E = \{P_t, P_f, p_{ROI}, \Sigma\}$, $S_R = \{j_{1,2}, p_{1,2}\}$; current action \stt{action}; request for context $req$
        \State \textbf{Output}: Set of context fluents $F$, target position $g$ for \stt{action} and failure flag $fail$
        \State \textbf{Init}: $fail = False$, $req = False$, \stt{action} = $Null$, $P_{close}\ =\ \{p \in P_t \ | \ ||p - p_{ROI}||_2<\SI{10}{\mm}\}$
        \Repeat
        \If{$req$}
            \State $F$ = compute\_fluents($S_E$, $S_R$, $P_{close}$)
        \EndIf
        \If{\stt{action} == \stt{pull}}
            \State $fail$ = check\_failure($S_E$)
        \EndIf
        \State $g$ = compute\_target(\stt{action})
        \Until{end of task}
    \end{algorithmic}
\end{algorithm}
\begin{algorithm}[t]
    \caption{compute\_fluents}\label{alg:fluents}
    \begin{algorithmic}[1]
        \State \textbf{Input}:$S_E = \{P_t, P_f, p_{ROI}, \Sigma\}$, $S_R = \{j_{1,2}, p_{1,2}\}$, $P_{close}$
        \State \textbf{Output}: Set of context fluents $F$
        \State \textbf{Init}: $F = \emptyset$, set $\delta \in [0,1]$
        \State Compute $P_v\ =\ \{p \in P_{close} \ | \ ||p - p_{ROI}||_2>\SI{10}{\mm}\}$
        \State \textit{\% ROI visibility ($|\cdot|$ is the cardinality of a set)}
        \If{$\frac{|P_v|}{|P_{close}|}>\delta$}
            \State $F$.append(\stt{visible\_ROI})
        \EndIf
        \State \textit{\% Gripper state}
        \For{$i \in \{1,2\}$}
            \If{$j_i <$\ang{20}}
                \State $F$.append(\stt{closed\_gripper(A$_i$)})
            \EndIf
        \EndFor
        \State \textit{\% Finite set of points}
        \State $B$ = discretize($P_t$)
        \For{$b \in B$}
            \State \textit{\% ROI location under tissue}
            \If{$p_{ROI} \in b$}
                \State $F$.append(\stt{above\_ROI(b)})
            \EndIf
            \State \textit{\% Reachability with respect to $x-$axis, see Sec. \ref{sec:task}}
            \State $j = argmin_{i=\{1,2\}} ||y_{i} - y_b||_1$
            \State $F$.append(\stt{reachable(A$_j$, b)})
            \State \textit{\% Points close to APs}
            \If{$\exists p \in P_f :\  p \in B$}
                \State $F$.append(\stt{fixed(b)})
            \EndIf
            \For{$i \in \{1,2\}$}
                \State \textit{\% Arm location}
                \If{$||p_i - b||_2 <$\SI{5}{\mm}}
                    \State $F$.append(\stt{at(A$_i$, b)})
                    \State \textit{\% Tissue grasped}
                    \If{$j_i <$\ang{20}}
                        \State $F$.append(\stt{in\_hand(A$_i$, b)})
                    \EndIf
                \EndIf
            \EndFor
        \EndFor
        \Return $F$
    \end{algorithmic}
\end{algorithm}
The SA module receives the state of the robot $S_R$, i.e. the 3D position of PSMs $p_{1,2}$ and joint angles of grippers $j_{1,2}$, from robot kinematics; and the state of the environment $S_E$, including the point cloud of tissue points $P_{t}$ and APs $P_f \subseteq P_t$; the set $\Sigma$ of mechanical forces at each tissue point from the biomechanical simulation; and the position of the ROI $p_{ROI}$. All the quantities in $S_R$ and $S_E$ are referred to the same kinematic frame (Fig.~\ref{fig:setup}). 
External fluents for task reasoning are computed by the function $compute\_fluents$ as described by Alg.~\ref{alg:fluents} (time \stt{t} is omitted because it is assigned at ASP level). 
Specifically, Lines 7-8 compute ROI visibility. To this purpose, at the beginning of the task (Line~3 in Alg.~\ref{alg:sa}) we compute $P_{close}$ as the set of tissue points close to the ROI within a fixed threshold (\SI{10}{\mm} in this work). The ROI is assumed to be visible when a pre-defined percentage (above a threshold $\delta$) of $P_{close}$ becomes far enough from the ROI itself (e.g., more than $10$\,mm in this work), thus indicating that the tissue has been displaced such that the ROI is accessible.
When an action has to be executed, function $compute\_target$ computes the motion target $g$ for it.
The target is needed only for actions \stt{reach(A, B, t)} (the position of grasping point \stt{B}, see Fig.~\ref{fig:setup}) and \stt{move(A$_i$, B, t)} ($(x_B, y_B, z_i)$). 
During the execution of \stt{pull(A, B, t)} action, the function $check\_failure$ verifies that $max\Sigma < \epsilon$, where $\epsilon$ is a threshold depending on the tissue mechanical properties. Otherwise, failure is notified to the MC module to interrupt the execution, and the task reasoner re-plans (with grounding of \stt{max\_height(A, t)}).

\subsubsection{Motion control}
The MC module receives the current action in the plan from the task reasoner. For each action, it selects the appropriate motion primitive from a pre-defined set, as in \cite{ginesi2020autonomous}. 
For actions \stt{reach(A, B, t)} and \stt{move(A, B, t)}, it also receives the goal position from the SA module. 
Otherwise, the target is ($x_i, y_i, z_i + h$) for \stt{pull(A$_i$, B, t)}, i.e. the tissue is pulled up to a pre-defined height ($h = $\SI{50}{\mm} in our task). Grasping and releasing of the gripper are direct joint closing and opening commands, respectively. 
In this work, we consider only linear trajectories, keeping a fixed orientation at the end-effector.

\subsubsection{Simulated environment}
A physics-based model of the anatomical environment is created starting from patient-specific geometry and mechanical properties extracted from pre-operative data. 
At each time instant, the simulation updates the tissue state based on the current robot configuration received from the MC module. To obtain a physically accurate result, we model the deformable behavior of anatomical tissues exploiting continuum mechanics laws, solved with the finite element (FE) method \cite{zhang2017deformable}. The choice of a physics-based method allows to continuously estimate forces exerted on the tissue, a crucial quantity to monitor and avoid unnecessary tissue damage. 
In the considered simulation environment, grasping is implemented by constraining the tissue points closest to the simulated PSM (distance below a small threshold, $5$\,mm in our task) to follow PSM motion, similarly to \cite{patil2010toward,tagliabue2020soft}. Grasping is activated or released based on commands received from the MC module. Current implementation does not consider the possibility of slippage or breaking contacts between the tool and the tissue, that might happen during real retraction. 


\section{RESULTS AND DISCUSSION}
\label{sec:exp}
For the experimental evaluation of FRAS, we use
the dVRK to control the surgical robot and receive kinematic information. 
The task reasoning module is implemented using
the state-of-the-art ASP solver Clingo \cite{gebser2008user}.
We assess performance of the framework on TR of a rectangular soft tissue sample of dimension $100\times120\times5$\,mm, a realistic size for robot-assisted surgical applications \cite{bhayani2008vinci}, in a simulated environment.
Hence, in this evaluation, all the information about the state of the environment $S_E$ (including the point cloud of tissue points $P_{t}$) are provided to the SA module by the simulation itself (Fig.~\ref{fig:framework}).
The tissue is modelled as a St Venant-Kirchhoff material with Young's modulus $3$\,kPa and Poisson ratio $0.45$, a common modelling choice for adipose tissues \cite{alkhouli2013mechanical,tagliabue2021data}, which are often responsible of hiding structures of interest during real surgery \cite{bhayani2008vinci}. APs are defined as random tissue patches, following the strategy proposed in \cite{tagliabue2021data}. 
In order to ensure computational tractability of selecting the grasping point at the logic reasoning level, we consider a finite set of candidate grasping points, uniformly spaced on an $N \times N$ grid, centered on the tissue of interest (Fig.~\ref{fig:setup}).
In all tests, we set the threshold for visibility $\delta = 70\%$ in Alg.~\ref{alg:fluents}.
Tests are run on a PC with 2.6 GHz Intel Core i7 processor and 16 GB RAM.



\subsection{Autonomous tissue retraction}
\begin{figure*}[t]
\centering
\begin{subfigure}{.24\linewidth}
  \centering
  \includegraphics[trim={150 350 150 200},clip,width=\linewidth]{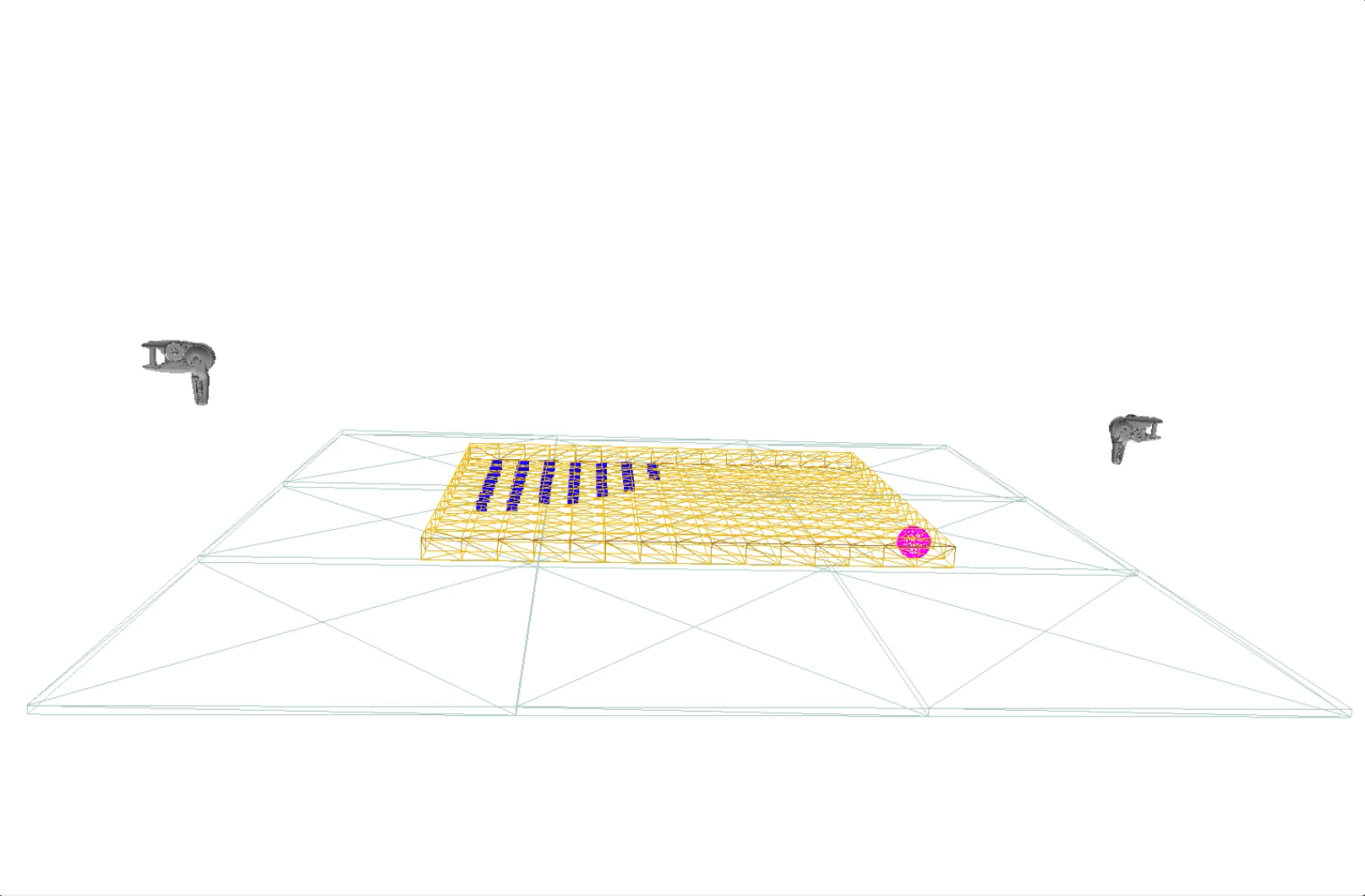}  
  \caption{Initial state}
\end{subfigure}
\begin{subfigure}{.24\linewidth}
  \centering
  \includegraphics[trim={150 350 150 200},clip,width=\linewidth]{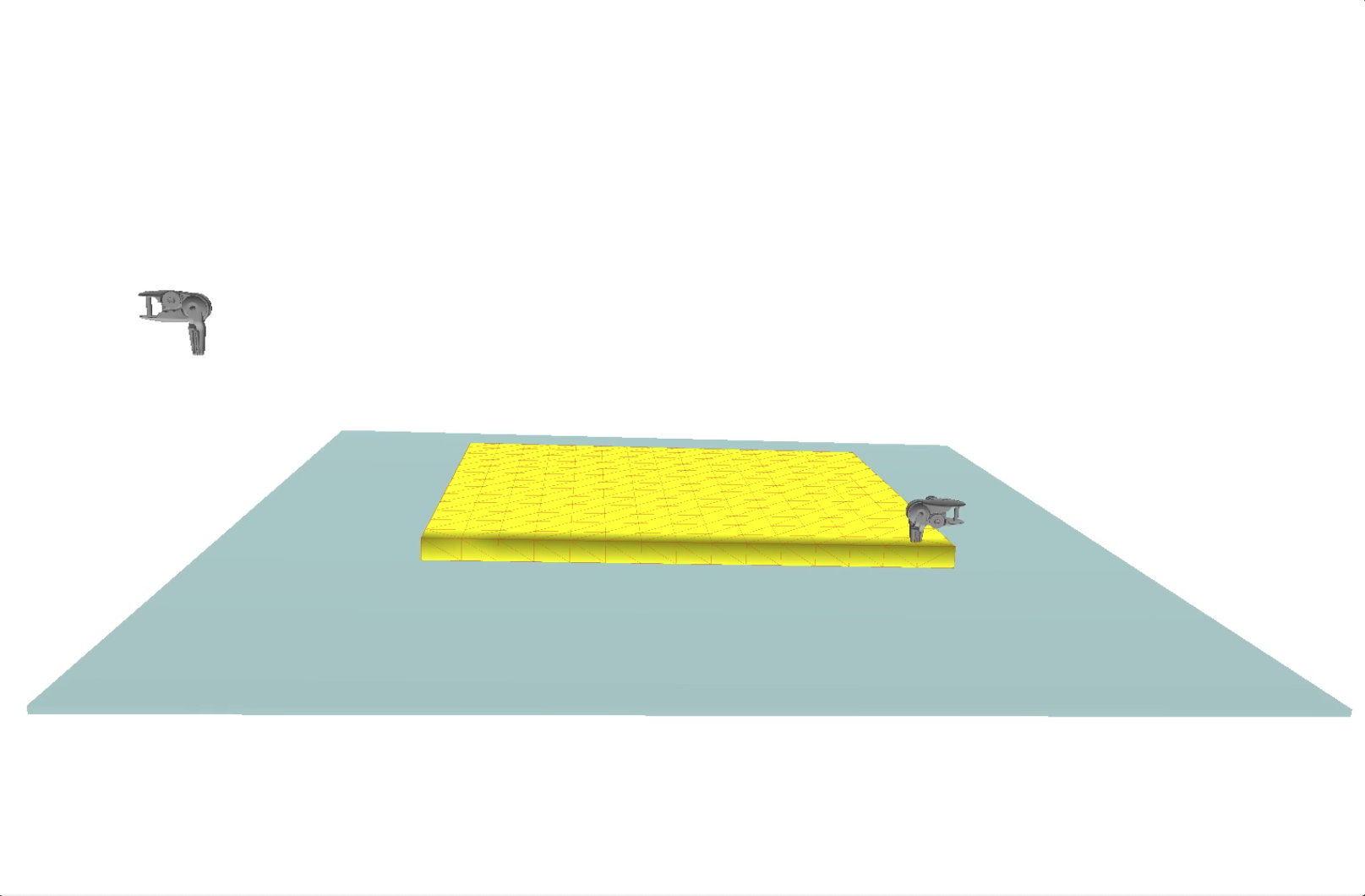}  
  \caption{Grasping tissue}
\end{subfigure}
\begin{subfigure}{.24\linewidth}
  \centering
  \includegraphics[trim={150 350 150 200},clip,width=\linewidth]{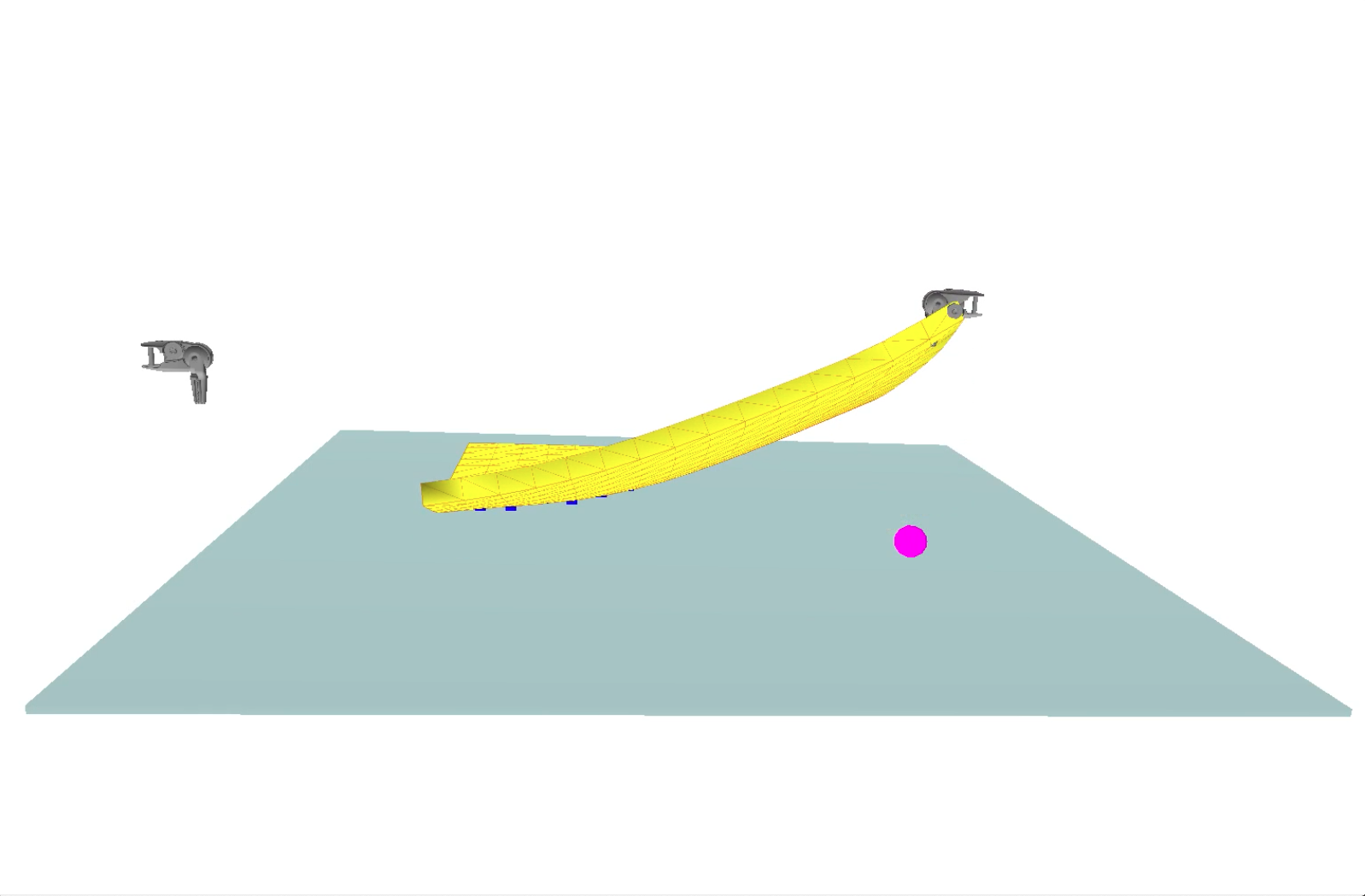}  
  \caption{ROI revealed}
\end{subfigure}
\caption{Scenario A, successful pulling with no replanning.}
\label{fig:sceneA}
\end{figure*}
\begin{figure*}[t]
\centering
\begin{subfigure}{.24\linewidth}
  \centering
  \includegraphics[trim={150 350 150 200},clip,width=\linewidth]{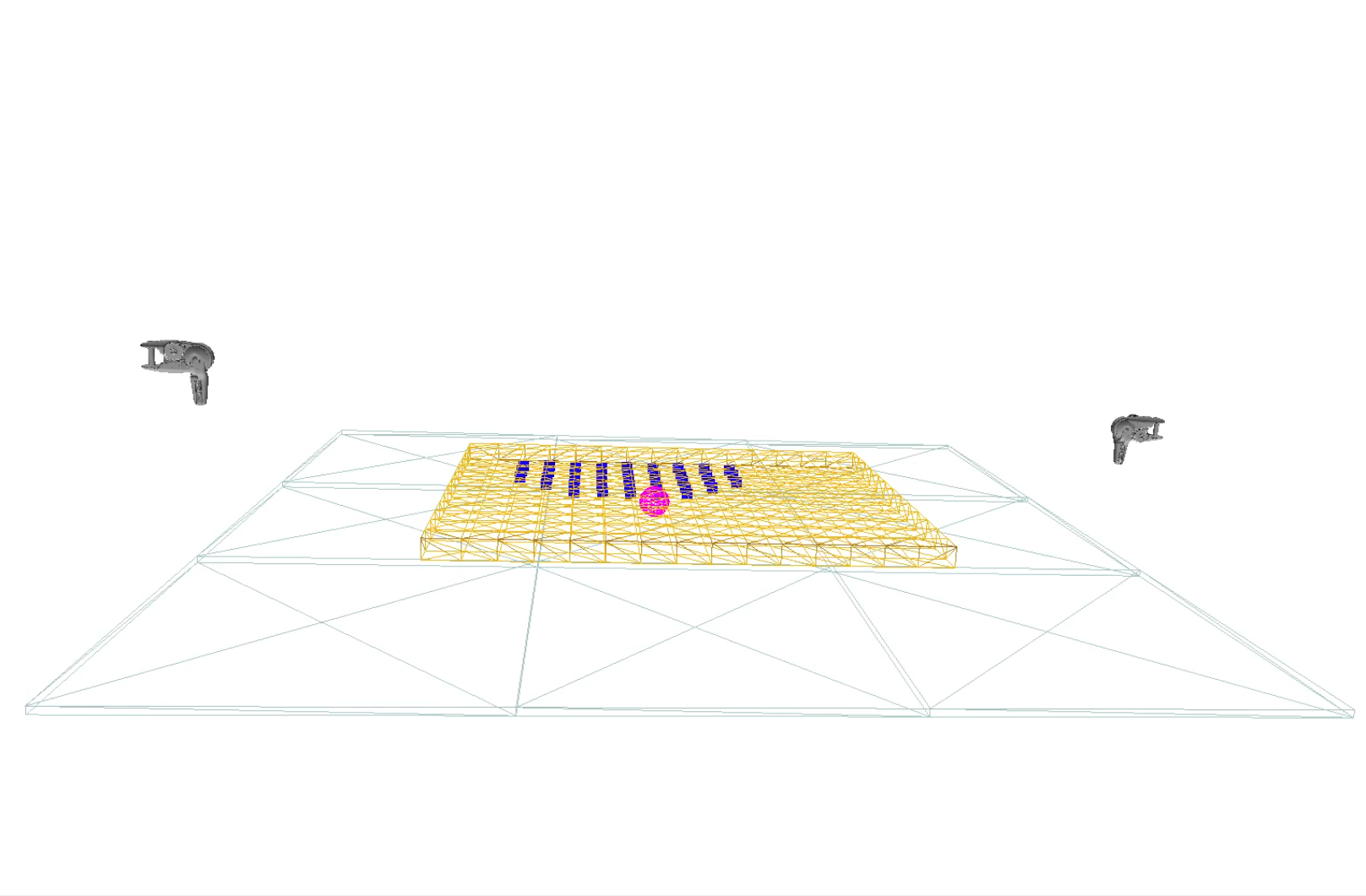}  
  \caption{Initial state}
\end{subfigure}
\begin{subfigure}{.24\linewidth}
  \centering
  \includegraphics[trim={150 350 150 200},clip,width=\linewidth]{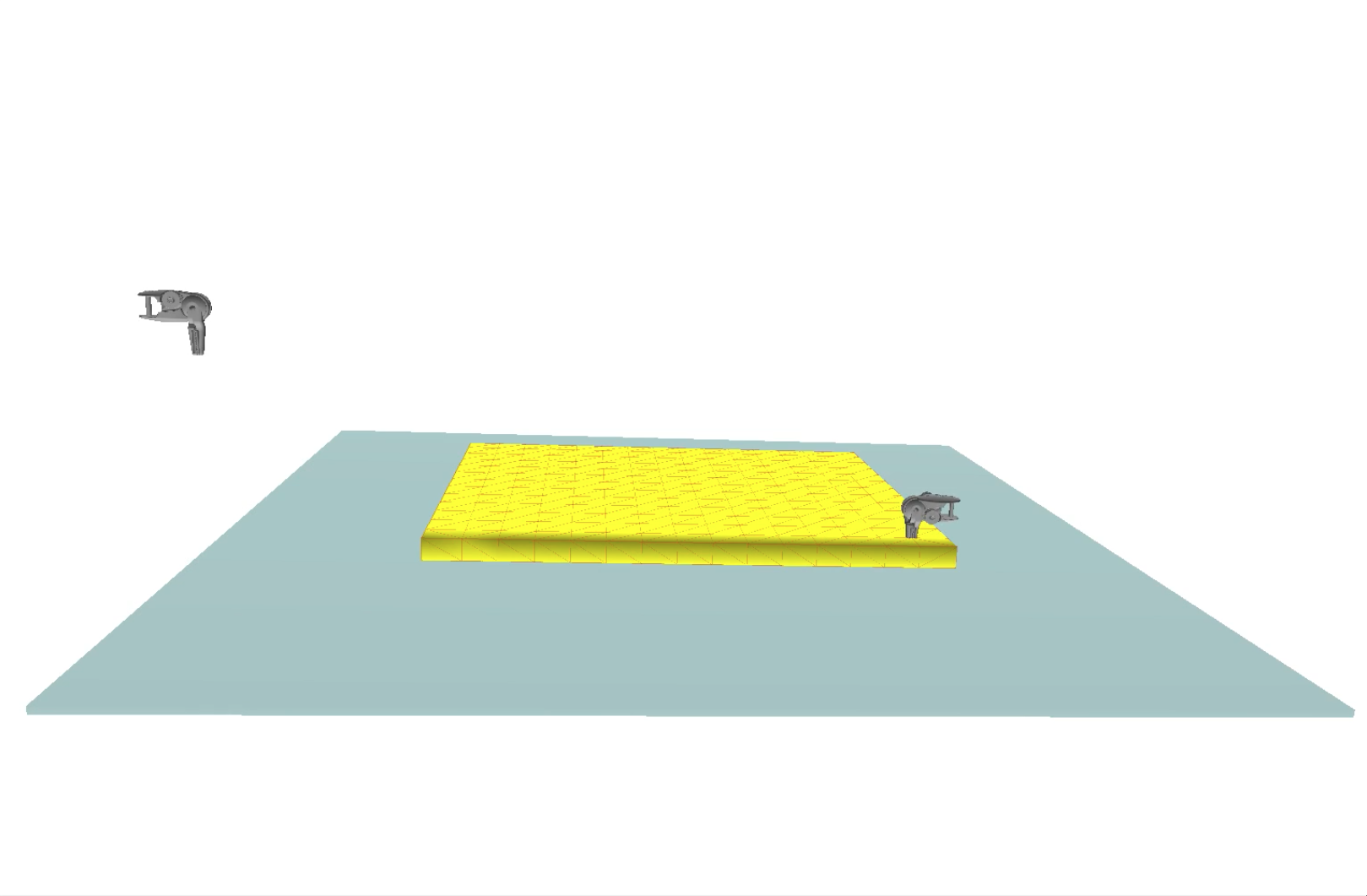}  
  \caption{Grasping tissue}
\end{subfigure}
\begin{subfigure}{.24\linewidth}
  \centering
  \includegraphics[trim={150 350 150 200},clip,width=\linewidth]{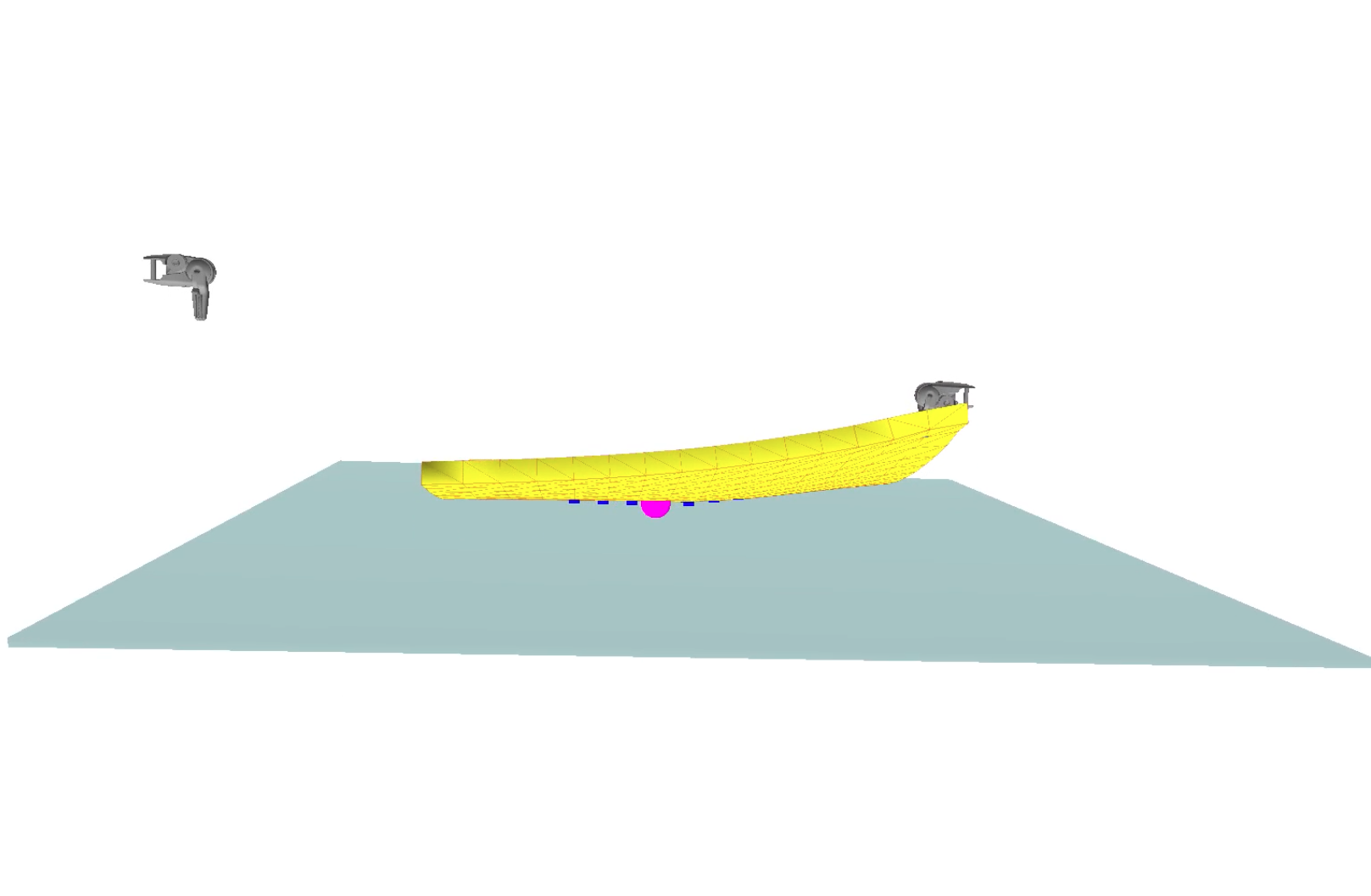}  
  \caption{Pulling interrupted}
\end{subfigure}
\begin{subfigure}{.24\linewidth}
  \centering
  \includegraphics[trim={150 350 150 200},clip,width=\linewidth]{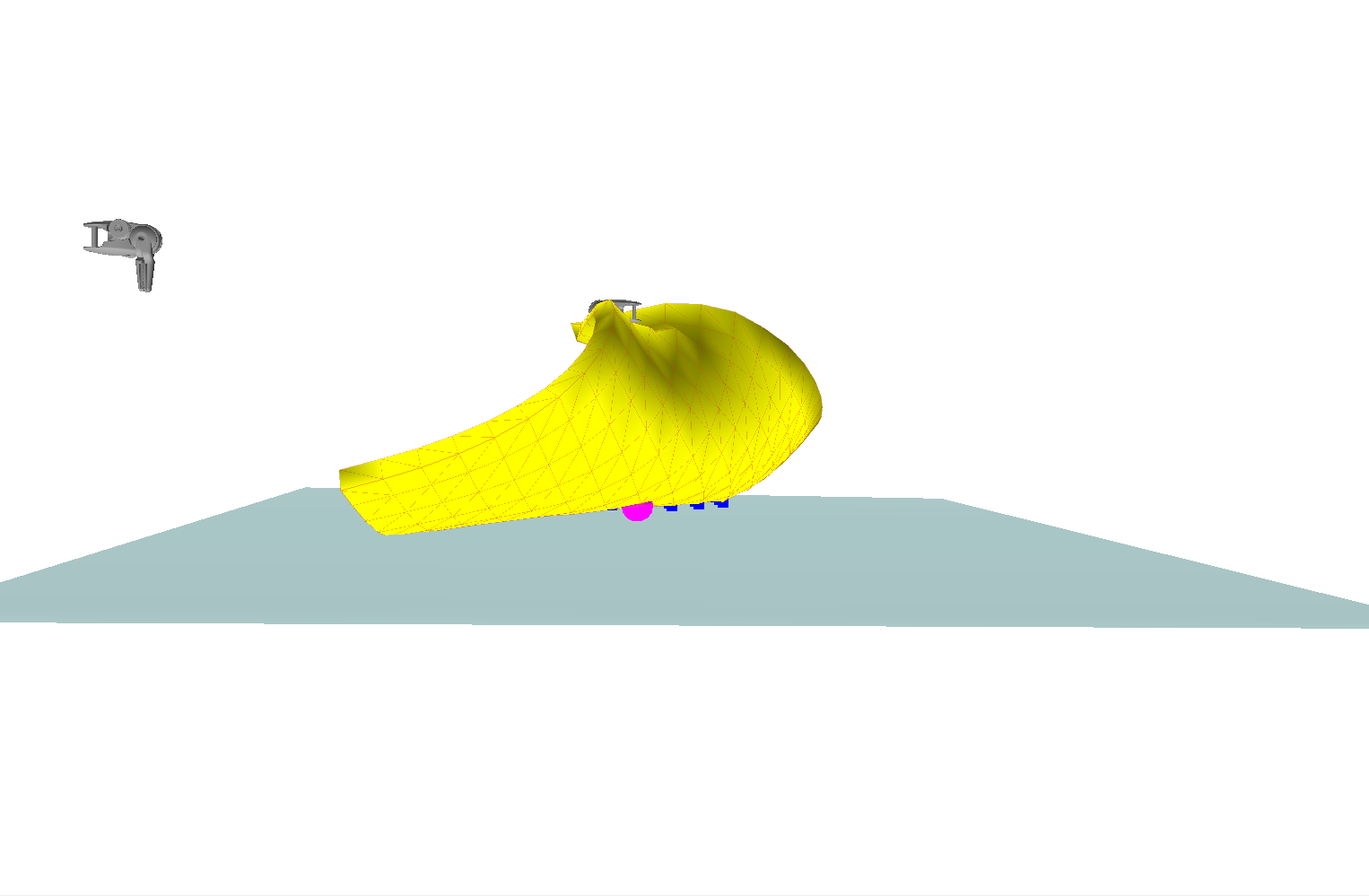}  
  \caption{ROI revealed}
\end{subfigure}
\caption{Scenario B, re-planning with \stt{move(A, B, t)} in c).}
\label{fig:sceneB}
\end{figure*}
\begin{figure*}[t]
\centering
\begin{subfigure}{.24\linewidth}
  \centering
  \includegraphics[trim={150 350 150 200},clip,width=\linewidth]{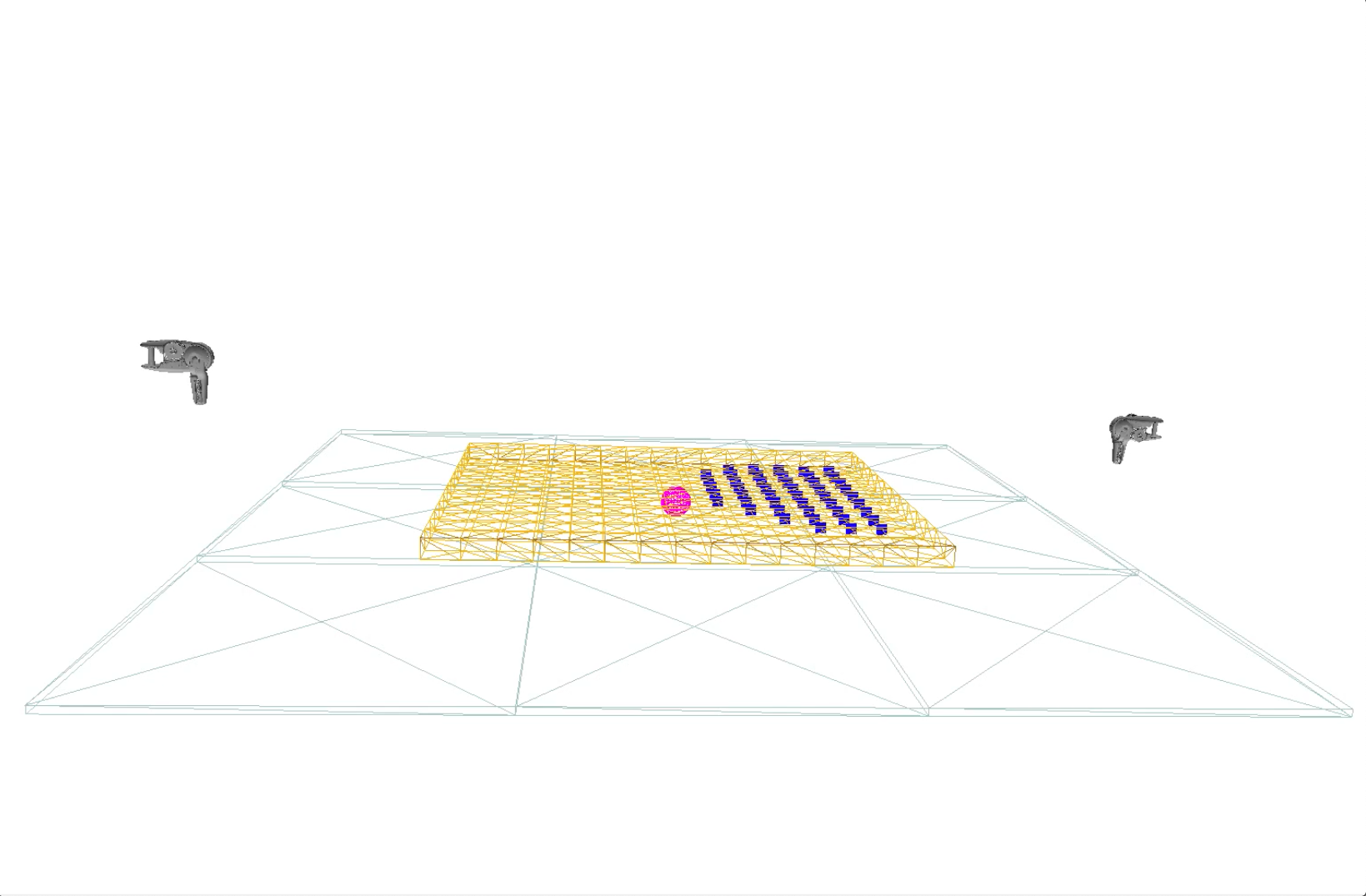}  
  \caption{Initial state}
\end{subfigure}
\begin{subfigure}{.24\linewidth}
  \centering
  \includegraphics[trim={150 350 150 200},clip,width=\linewidth]{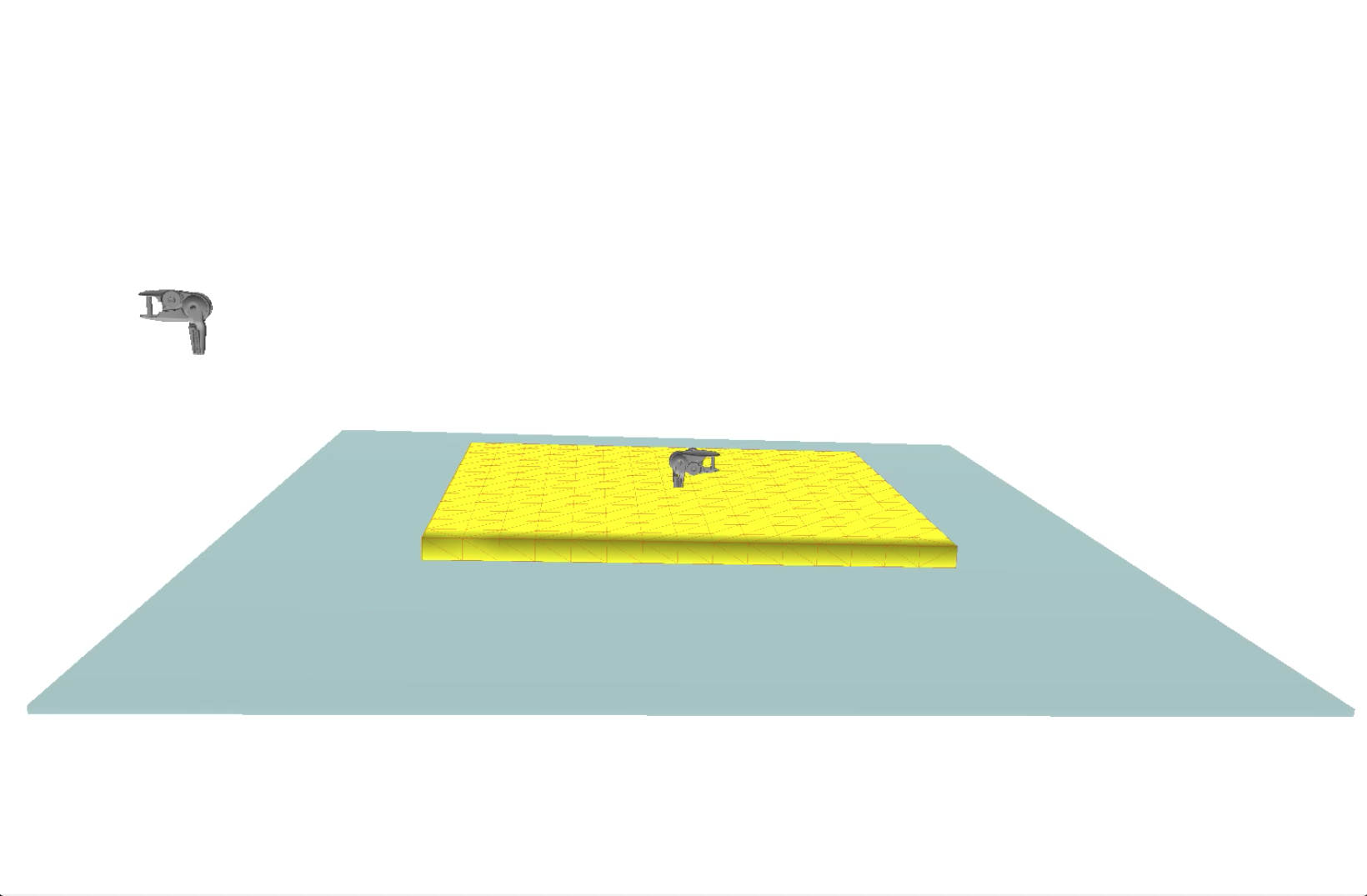}  
  \caption{First grasping}
\end{subfigure}
\begin{subfigure}{.24\linewidth}
  \centering
  \includegraphics[trim={150 350 150 200},clip,width=\linewidth]{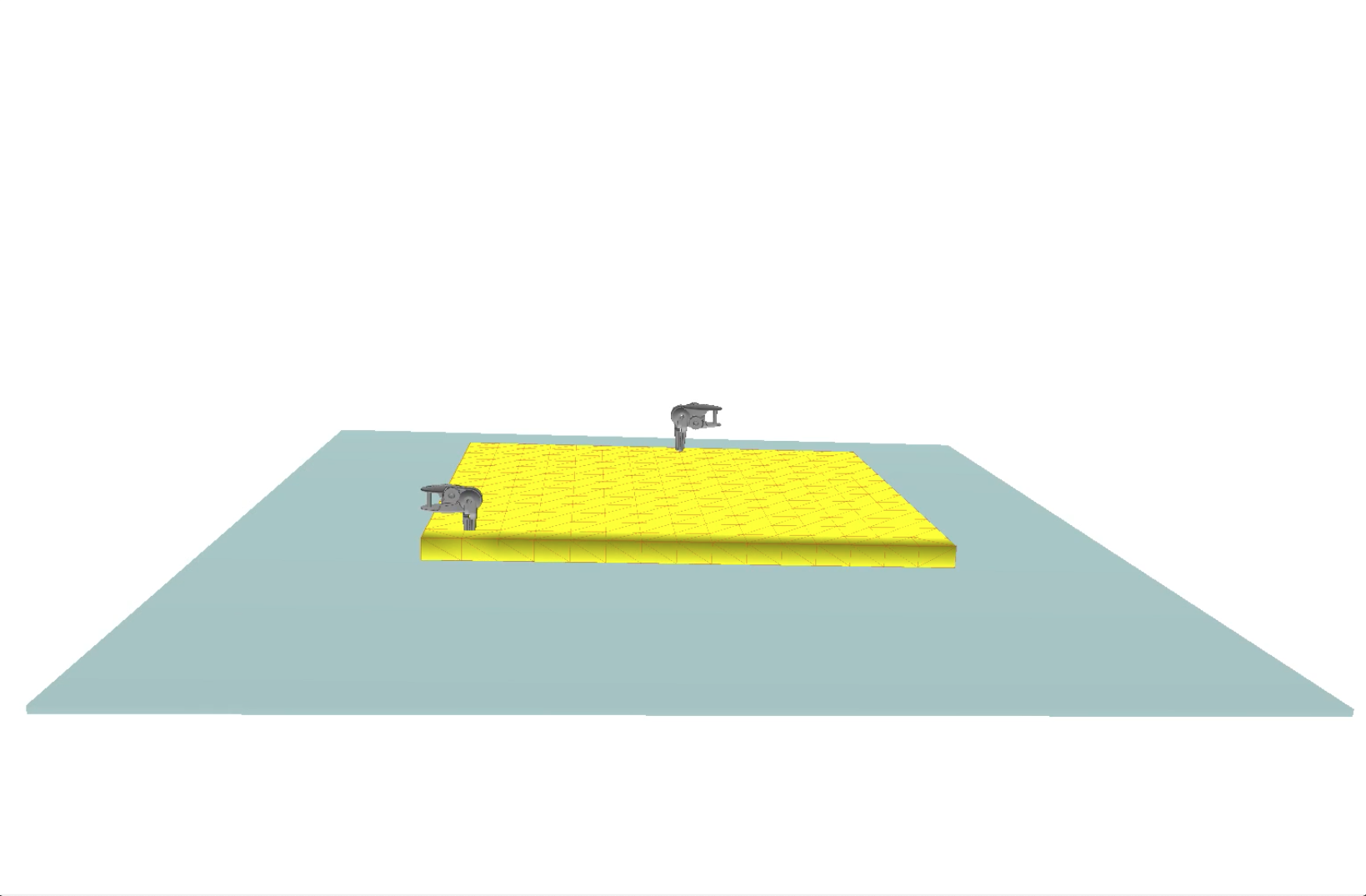}  
  \caption{Second grasping}
\end{subfigure}
\begin{subfigure}{.24\linewidth}
  \centering
  \includegraphics[trim={150 350 150 200},clip,width=\linewidth]{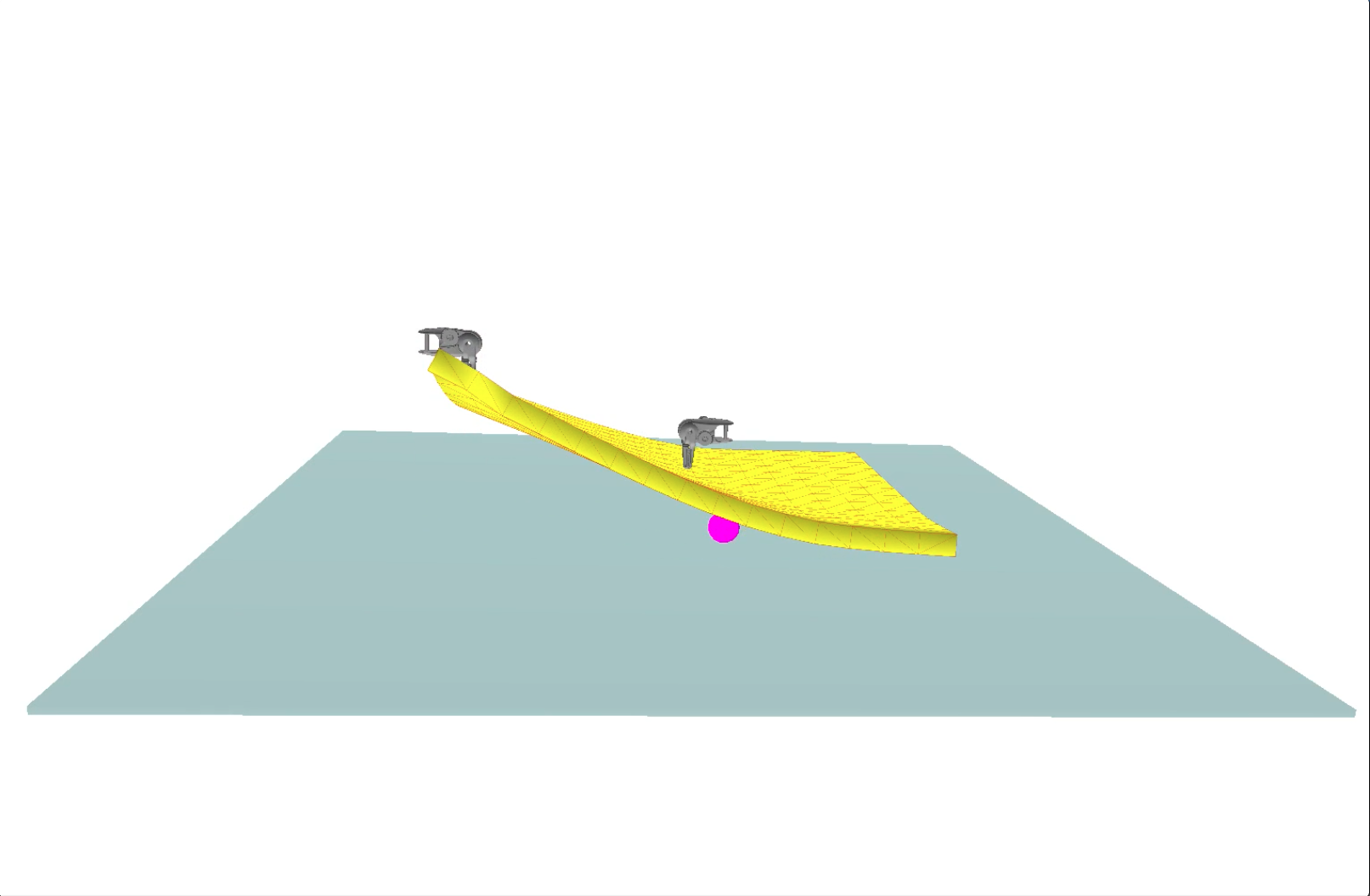}  
  \caption{ROI revealed}
\end{subfigure}
\caption{Scenario C, re-planning after grasping close to an attachment point in b).}
\label{fig:sceneC}
\end{figure*}

We run experiments to evaluate the capability of the framework to perform autonomous TR handling the different conditions that may occur. 
In this evaluation, we consider candidate grasping points distributed on a grid of resolution $7 \times 7$ (i.e., one possible grasping point every $15$\,mm).
The force limit is set to $\epsilon =$ \SI{0.5}{\N}, to avoid any possible dangerous tissue manipulation.
Success rate of FRAS is assessed by measuring the average visibility percentage as defined in Alg.~\ref{alg:sa}, reached upon task completion.

We run 100 simulations initialized with different random environmental states (i.e., $p_{ROI}$ and $P_f$). After the tissue is grasped, two possible situations might happen, depending solely on the initial state. In the simplest case (Scenario A in Fig.~\ref{fig:sceneA}), the force applied to the tissue never exceeds the defined threshold $\epsilon$; thus, the tissue is pulled upwards until the ROI is exposed. Instead, in case the force overcomes the safety threshold $\epsilon$, the PSM moves the tissue laterally towards the ROI in order to reveal it (Scenario B in Fig.~\ref{fig:sceneB}). 

To better assess the contribution of the force feedback, we consider a third situation where the same 100 simulations are run again, but ignoring the constraint on the distance from APs for grasping point selection (Scenario C in Fig.~\ref{fig:sceneC}). In this situation, the grasping point is selected as the one closest to the ROI, regardless of the position of APs. 
This scenario allows to evaluate how FRAS would react in case the chosen grasping point is very close to an AP, which may happen when there are many APs and the ROI is very close to them. 
We note that when grasping close to the APs, the force goes beyond $\epsilon$ immediately after grasping (i.e., when the $z-$position of the PSM is close to the tissue). Then, the task reasoner must re-plan grasping to another point, now maximizing also the distance from APs.

The three scenarios are illustrated in the video attachment to this paper, evidencing the interpretability of the plan and the environmental context description. Even though a simplified tissue model is considered for evaluation in this paper, it is worth noticing that our framework can provide a plan for any tissue geometry and state of the environment.

We have found that re-planning with Scenario B occurs in 36 simulations, while re-planning in Scenario C is needed in 46 cases. 
The achieved visibility percentage (mean $\pm$ standard deviation) is $96.89\% \pm 6.37\%$ (Scenarios A, B and C), demonstrating that, on average, our autonomous framework is able to expose the region of interest to a high extent, thus successfully accomplishing the task. Results show that re-planning does not affect the final visibility percentage (average visibility is $96.41\%$ when re-planning is needed in Scenarios B and C, comparable to the average over all executions).

One of the contributions of this work lies in the integration of the dynamic feedback from the biomechanical simulation for enhanced situation awareness. Our results demonstrate that the force feedback provided by the simulation triggers re-planning in slightly more than one third of the cases. In real surgery, where the surgeon has no information about the force applied to the tissue, all these cases represent potentially dangerous situations where tissue damage is highly probable. 
In fact, the mechanical stress on the tissue can only be roughly evaluated from visual feedback or based on the surgeon's expertise. In particular, when grasping very close to APs (Scenario C), the force feedback allows to immediately recognize the high risk of tissue damage and adapt the behavior to successfully complete the task.

We further verify the importance of force feedback from simulation by running a set of 100 simulations where no force limit is set. In this batch of simulations, the maximum force applied to the tissue is $(1.25 \pm 0.96)$ \si{\N} (mean $\pm$ standard deviation), i.e. almost 3 times the pre-defined threshold, with a peak of \SI{8.50}{\N} (when grasping occurs close to APs). 
As a consequence, integration of force information from the simulation is crucial to minimize the risk of tissue tearing or rupture.

\subsection{Planning efficiency}
An additional requirement of the task reasoning module is responsive reaction to the quickly changing anatomical environment. 
The computational performance of task reasoner is evaluated as the \emph{planning time}, defined as the time required by Clingo to compute a plan, given a context interpretation from the SA module. 
Hence, it does not depend on the rate at which sensor data are received.
The planning time includes time for grasping point selection; thus, it may be affected by two main parameters: the number of candidate grasping points and the percentage $f_\%$ of APs. In fact, they both influence the size of the search space for possible grasping points.
In this evaluation, we consider different resolutions of the grid defining candidate grasping points $N \in \{5, ..., 10\}$. For each value of $N$, we run 25 simulations for each value $f_\% \in \{10\%, 30\%, 50\%, 70\%\}$. This results in 600 total simulations, each initialized with a different random $p_{ROI}$ and $P_f$ (according to the value of $f_\%$). 

\begin{table*}[t]
\caption{Planning time and re-planning times in Scenarios B and C (average $\pm$ standard deviation), with different tissue grid size $N$ and percentage of APs $f_\%$.}
\label{tab:times}
\begin{center}
  \begin{tabular}{c|c|c|c|c|c|c}
    \multirow{2}{*}{$\mathbf{N}$} &
      \multicolumn{4}{c|}{\textbf{Planning time [\si{\s}]}} &
      \multicolumn{1}{c|}{\textbf{Re-planning time B [\si{\s}]}} &
      \multicolumn{1}{c}{\textbf{Re-planning time C [\si{\s}]}} \\
      \cline{2-7}
    & $\mathbf{f_\% = 10\%}$ & $\mathbf{f_\% = 30\%}$ & $\mathbf{f_\% = 50\%}$ & $\mathbf{f_\% = 70\%}$ & $\mathbf{f_\% = 50\%}$ & $\mathbf{f_\% = 50\%}$  \\
    \hline
    $\mathbf{5}$ & $0.25 \pm 0.02$ & $0.24 \pm 0.02$ & $0.23 \pm 0.01$ & $0.24 \pm 0.01$ & $0.12 \pm 0.04$ & $0.32 \pm 0.02$ \\
    \hline
    $\mathbf{6}$ & $0.68 \pm 0.02$ & $0.65 \pm 0.02$ & $0.63 \pm 0.01$ & $0.62 \pm 0.01$ & $0.32 \pm 0.09$ & $1.01 \pm 0.10$ \\
    \hline
    $\mathbf{7}$ & $1.50 \pm 0.02$ & $1.48 \pm 0.03$ & $1.41 \pm 0.02$ & $1.42 \pm 0.01$ & $0.72 \pm 0.19$ & $1.95 \pm 0.04$ \\
    \hline
    $\mathbf{8}$ & $3.06 \pm 0.04$ & $2.97 \pm 0.06$ & $2.90 \pm 0.04$ & $2.88 \pm 0.02$ & $1.52 \pm 0.49$ & $4.03 \pm 0.08$ \\
    \hline
    $\mathbf{9}$ & $5.88 \pm 0.08$ & $5.70 \pm 0.15$ & $5.51 \pm 0.03$ & $5.51 \pm 0.05$ & $2.77 \pm 0.60$ & $7.74 \pm 0.16$ \\
    \hline
    $\mathbf{10}$ & $10.75 \pm 0.20$ & $10.21 \pm 0.28$ & $9.92 \pm 0.09$ & $9.87 \pm 0.08$ & $4.95 \pm 1.05$ & $16.00 \pm 1.43$ \\
    \hline
  \end{tabular}
 \end{center}
\end{table*}
The second column of Table~\ref{tab:times} reports the time required by the task reasoner to generate the initial plan.
Obtained results show that the most influencing parameter is $N$, while $f_\%$ does not significantly affect the average planning time.
The task reasoner is able to compute the grasping point and a plan within \SI{1.5}{\s} when using a $7 \times 7$ grid discretization for grasping point definition, demonstrating high efficiency. 
Although here we do not intend to provide a strategy for optimal grasping point selection, obtained results suggest that the proposed approach can be used in combination with state-of-the-art grasping point selection algorithms, to significantly improve their performance through search space shrinkage even in real surgery \cite{jansen2009surgical}.

The computational capabilities of FRAS are also validated in terms of re-planning time (Scenarios B-C). 
Results are reported in the last two columns of Table~\ref{tab:times}.
We run 100 simulations for each scenario and for each value of $N$, setting constant $f_\% = 50 \%$ (since $f_\%$ has shown not to influence the planning time). 
The average re-planning time for Scenario B is approximately half of the planning time obtained at the same value of $N$. This happens because the initial grasping point is far from APs, hence the system does not need to select another grasping point, which represents a significant part of the ASP computational effort.
On the contrary, in Scenario C the system requires more time to re-plan because tissue releasing and new grasping point selection are required.
Overall, the re-planning time is below \SI{2}{\s} for a $7 \times 7$ grasping points discretization, which guarantees efficient adaptation to the quickly changing anatomical environment, introducing only a minor delay which is compatible with intra-operative times.

\section{CONCLUSIONS}
\label{sec:conclusion}
This work presents FRAS, a logic-based framework for autonomous surgical task execution, supporting real-time planning and failure recovery in deformable anatomical environment. Performance of FRAS is preliminarly assessed on a simulated tissue retraction task.
Logic-based task reasoning and real-time situation awareness guarantee interpretability of the workflow of execution (environmental context interpretation and consequent plan generation), which is necessary for monitoring and to guarantee the reliability of the ARSS.
Logics can also reduce the computational burden of standard state-of-the-art algorithms for grasping selection.
A biomechanical simulation provides feedback on the forces applied on the tissue, reducing the risk of tissue damage.
Future works will focus on the validation on a real scenario, where the role of the SA module and the simulation will be fundamental to close the control loop and ensure reliable execution.
Furthermore, we plan to exploit the output from the SA module to continuously update the simulated environment and compensate for possible modelling uncertainties \cite{wu2020leveraging}.
In particular, we will integrate strategies to update APs during interaction with the tissue, such as \cite{nikolaev2020estimation,tagliabue2021intra}, compensating for partial or imprecise knowledge about their position at the beginning of the task.
Extension to more complex surgical tasks would also require the integration with more sophisticated motion planning strategies \cite{ginesi2020autonomous}.



\bibliographystyle{./IEEEtran}
\bibliography{./IEEEabrv,./bibliography}

\end{document}